\documentclass[10pt,leqno]{amsart}
\usepackage{graphicx}
\usepackage{indentfirst,csquotes}

\usepackage{amssymb,amsthm,amsmath}
\usepackage{xcolor,paralist,hyperref,fancyhdr,etoolbox}
\usepackage{booktabs}
\usepackage{listings}
\usepackage{microtype}
\usepackage{array}

\hypersetup{ colorlinks=true, linkcolor=black, filecolor=black, urlcolor=black }

\lstdefinelanguage{Python}{
  keywords={import,from,class,def,return,if,else,for,in,not,is,None,True,False,with,as,assert,raise,try,except,finally,lambda,yield,pass,break,continue,global,nonlocal,async,await},
  keywordstyle=\color{blue!70!black}\bfseries,
  ndkeywords={self,cls,Tensor,SafetyState,SafetyFilterResult,ObsT,ActT},
  ndkeywordstyle=\color{purple!70!black},
  commentstyle=\color{gray}\itshape,
  stringstyle=\color{teal!70!black},
  basicstyle=\ttfamily\small,
  showstringspaces=false,
  breaklines=true,
  numbers=none,
  frame=single,
  framerule=0.4pt,
  rulecolor=\color{gray!50},
  xleftmargin=8pt,
  xrightmargin=8pt,
}

\begin{document}

\title[ParallelCBF]{ParallelCBF: A Composable Safety-Filter and Auditability Framework for Tensor-Parallel Reinforcement Learning}

\author[Y.~Ma]{Yuyin Ma}
\address{Xinjiang Key Laboratory of Intelligent Computing and Smart Applications, School of Software, Xinjiang University, Urumqi, China.}
\email{mayuyin@xju.edu.cn}

\author[Z.~Yang]{Zilei Yang}
\address{Xinjiang Key Laboratory of Intelligent Computing and Smart Applications, School of Software, Xinjiang University, Urumqi, China.}
\email{107552504876@stu.xju.edu.cn}

\author[Y.~Lu]{Yijun Lu}
\address{School of Computing Science, Waseda University, Tokyo, Japan.}
\email{yijun@ruri.waseda.jp}

\date{\today}
\maketitle

\let\thefootnote\relax
\footnotetext{Keywords: Control Barrier Functions; Safe Reinforcement Learning; Tensor-Parallel Computing; Reproducibility; Open-Source Software.}

\begin{abstract}
While Isaac Lab provides massive parallel UAV simulation, OmniSafe and
safe-control-gym provide constrained-RL benchmarks, and CBFKit provides
control-barrier-function synthesis tooling, no existing framework unifies
these capabilities for end-to-end safety-constrained training.
ParallelCBF is the first framework to unify (i)~tensor-parallel UAV
environments, (ii)~hard-gate CBF safety filters, (iii)~sharded BC-to-RL
pipelines, and (iv)~first-class operational auditability---pre-registration,
watchdog registries, failure forensics, and dataset audits as composable
APIs rather than user-implemented scripts. We release ParallelCBF v0.1.0
under Apache~2.0 with a four-layer composable API, a CPU PyTorch reference
implementation of a dual-barrier (squared / linear-predictive) CBF,
property-based safety invariance tests across vectorized batch sizes that
complete in 1.67~s for the full 39-test suite, and a 31{,}415-episode
behavior-cloning collection campaign whose curriculum mix, per-bucket
yields, and dataset SHA-256 are auditable through the framework's own
\texttt{ops} primitives. We report a representative end-to-end pipeline
execution in which the framework's auditability layer halted a downstream
training stage that did not meet pre-registered convergence criteria,
preventing silent propagation of a degraded checkpoint---an architectural
property we argue is necessary, not merely useful, for reproducible
empirical robotics research. The framework is installable via
\texttt{pip install parallelcbf}; source and release artifacts are
available at \url{https://github.com/xiaoyang-123-cell/ParallelCBF}.
\end{abstract}

\bigskip

\section{Introduction}
\label{sec:intro}

Safety-critical reinforcement learning at scale faces a structural
fragmentation problem. Tensor-parallel simulators~\cite{isaaclab} solve
throughput. Constrained-RL libraries~\cite{omnisafe,safetygym} solve
algorithmic objectives. Control-theoretic toolkits~\cite{cbfkit} solve
barrier-function synthesis. Yet a researcher building a safety-constrained
UAV training pipeline must compose these artifacts manually, with no
shared abstraction for the safety filter, no standard interface for
behavior-cloning-to-RL warmstart, and no framework-level support for the
operational disciplines---pre-registration, watchdog logic, failure-mode
capture---that distinguish reproducible empirical work from one-off
training runs.

This paper presents \textbf{ParallelCBF}, a framework that addresses this
fragmentation through four explicit architectural commitments:

\begin{enumerate}
  \item A \texttt{SafetyFilter} abstraction (with a \texttt{SafetyWrapper}
    composition primitive) decoupling barrier-function evaluation from
    environment dynamics.
  \item A reference dual-barrier CBF implementation in PyTorch, supporting
    vectorized evaluation across $N$ parallel environments and verified
    by property-based tests of the positive-invariance safety property.
  \item Tensor-parallel safe-RL pipeline primitives (sharded BPTT-aware
    dataloaders, BC checkpointing, KL-anchored algorithm interfaces)
    sufficient to bridge a behavior-cloning warmstart with downstream RL
    fine-tuning.
  \item A first-class \emph{operational auditability} layer
    (\texttt{parallelcbf.ops}) exposing pre-registration, watchdog
    registries, failure forensics, atomic checkpoints, and dataset
    audits as composable APIs.
\end{enumerate}

The fourth contribution is the most architecturally novel: to our
knowledge, ParallelCBF is the first robotics RL framework to elevate
operational concerns from user-implemented infrastructure to versioned,
type-checked, test-covered API surfaces. We argue this elevation is
necessary---not merely useful---for the field's
reproducibility properties~\cite{henderson2018deep,pineau2021reproducibility},
and we present the framework's first public release (v0.1.0)
together with a 31{,}415-episode behavior-cloning case study
(Section~\ref{sec:empirical}) whose dataset is anchored by a SHA-256
commitment recorded in the release manifest, as empirical demonstration.

\paragraph{Scope of v0.1.0.}
This release is deliberately narrow: pure Python, NumPy, and CPU PyTorch
references; a 2D toy environment; and the full auditability layer.
Isaac Lab adapters, Mamba-based policy backbones, and KL-anchored PPO
training are deferred to v0.2.0 (Phase~2 of the project roadmap),
following the principle that the framework's contracts should be
validated on toy reference workloads before being committed to
multi-million-step training campaigns.

\section{Related Work and Differentiation}
\label{sec:related}

We position ParallelCBF against four nearest-neighbor frameworks in the
safety-relevant robotics learning ecosystem (Table~\ref{tab:comparison}).

\textbf{Isaac Lab}~\cite{isaaclab} provides tensor-parallel simulation
infrastructure for robotics RL but exposes neither a safety-filter
abstraction nor an auditability layer; safety-constrained training on
Isaac Lab is currently the user's responsibility.

\textbf{OmniSafe}~\cite{omnisafe} and \textbf{Safety-Gymnasium}~\cite{safetygym}
provide constrained-RL algorithmic infrastructure (Lagrangian methods,
PCPO, projection-based algorithms) and benchmark environments, but do
not provide hard-gate CBF safety filters or framework-level auditability
primitives; safety is treated algorithmically through soft constraints
rather than as a structural commitment.

\textbf{safe-control-gym}~\cite{safecontrolgym} provides classical
control-theoretic baselines for safety-constrained learning, including
single-environment CBF examples, but does not extend to tensor-parallel
training nor expose auditability primitives.

\textbf{CBFKit}~\cite{cbfkit} provides control-barrier-function synthesis
and analysis tooling, primarily for theoretical study and controller
design, with no integration into RL training pipelines or
behavior-cloning workflows.

\begin{table}[t]
\centering
\scriptsize
\caption{Capability comparison across safety-relevant robotics learning
frameworks. Columns: \textbf{PCBF}~=~ParallelCBF, \textbf{IL}~=~Isaac Lab,
\textbf{OS}~=~OmniSafe, \textbf{SCG}~=~safe-control-gym,
\textbf{CBFK}~=~CBFKit. $\checkmark$~=~supported as a framework primitive;
$\times$~=~not provided; $\circ$~=~partial.}
\label{tab:comparison}
\begin{tabular}{@{}lccccc@{}}
\toprule
\textbf{Capability} & \textbf{PCBF} & \textbf{IL} & \textbf{OS} & \textbf{SCG} & \textbf{CBFK} \\
\midrule
Simulator-agnostic safety API     & $\checkmark$ & $\times$    & $\times$    & $\times$    & $\times$ \\
Composable safety wrapper         & $\checkmark$ & $\times$    & $\times$    & $\times$    & $\times$ \\
CBF safety-filter abstraction     & $\checkmark$ & $\times$    & $\times$    & $\times$    & $\checkmark$ \\
PyTorch batched CBF reference     & $\checkmark$ & $\times$    & $\times$    & $\times$    & $\times$ \\
Property tests for invariance     & $\checkmark$ & $\times$    & $\times$    & $\circ$     & $\circ$ \\
Parallel/vectorized fixture       & $\checkmark$ & $\checkmark$ & $\checkmark$ & $\times$    & $\times$ \\
\textbf{Watchdog registry}        & $\checkmark$ & $\times$    & $\times$    & $\times$    & $\times$ \\
\textbf{Failure-forensics buffer} & $\checkmark$ & $\times$    & $\times$    & $\times$    & $\times$ \\
\textbf{Atomic checkpoint helper} & $\checkmark$ & $\times$    & $\times$    & $\times$    & $\times$ \\
\textbf{Pre-registration commit}  & $\checkmark$ & $\times$    & $\times$    & $\times$    & $\times$ \\
\textbf{Dataset audit as plugin}  & $\checkmark$ & $\times$    & $\times$    & $\times$    & $\times$ \\
\bottomrule
\end{tabular}
\end{table}

The bolded rows---the auditability layer---are ParallelCBF's column-only
contributions and the architectural focus of this paper.

\section{Architecture}
\label{sec:arch}

\begin{figure}[t]
\centering
\includegraphics[width=\linewidth]{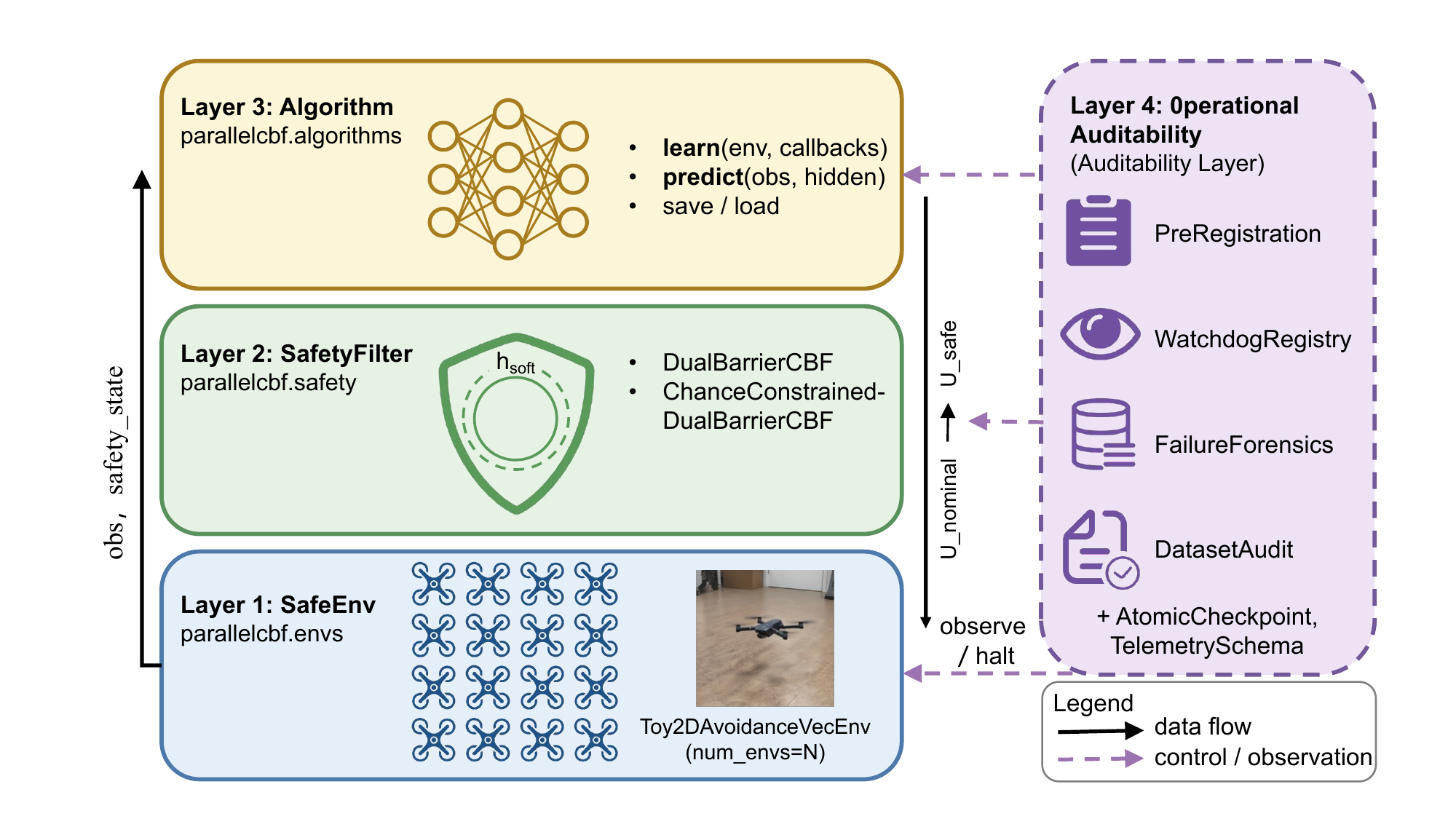}
\caption{ParallelCBF's four-layer architecture. From bottom to top:
tensor-parallel \texttt{SafeEnv} instances (Layer~1, the simulation
surface) drawn as a grid of replicated drone silhouettes; the
\texttt{SafetyFilter} abstraction with its dual-barrier safety shield
enveloping the central drone (Layer~2); the policy/algorithm composing
nominal actions, depicted as a data-flow constellation rising from the
shield (Layer~3); and the operational auditability primitives---seal,
watchdog sentinel, and forensics vault---suspended above the stack
(Layer~4), orthogonal to and overseeing the training pipeline.}
\label{fig:architecture}
\end{figure}

ParallelCBF organizes its public API into four explicit layers, each
defined by an abstract base class (ABC) and instantiated by reference
implementations. Figure~\ref{fig:architecture} provides a high-level
visual of this organization.

\subsection{Layer 1: Environments (\texttt{parallelcbf.envs})}

\texttt{SafeEnv} extends \texttt{gymnasium.Env} with three additional
contracts: \texttt{safety\_state()} returns a typed \texttt{SafetyState}
record (relative positions, velocities, drone state);
\texttt{safety\_metrics()} returns a structured metric dictionary;
\texttt{hard\_constraint\_violations()} returns a typed
\texttt{ConstraintViolations} record. The reference implementation
\texttt{Toy2DAvoidanceEnv} is a 2D point-mass with a single circular
obstacle; its vectorized variant \texttt{Toy2DAvoidanceVecEnv}
parameterizes over \texttt{num\_envs} for tensor-parallel rollouts.

\subsection{Layer 2: Safety Filters (\texttt{parallelcbf.safety})}

\texttt{SafetyFilter} is the framework's headline abstraction. Its
contract is:

\begin{lstlisting}[language=Python]
def filter_action(
    self,
    observation: ObsT,
    nominal_action: ActT,
    safety_state: SafetyState,
) -> SafetyFilterResult[ActT]: ...
\end{lstlisting}

The filter consumes the policy's nominal action, the current
\texttt{SafetyState}, and the observation (for learned filters that
condition on perception); it returns a \texttt{SafetyFilterResult}
containing the projected safe action, a diagnostic flag, and per-barrier
values.

Building on the standard control-barrier-function
framework~\cite{ames2019cbf}, the reference implementation
\texttt{DualBarrierCBF} provides a two-barrier formulation:
\begin{align}
h_{\text{hard}}(\mathbf{x})    &= \|\mathbf{r}\|^2 - R^2
  \qquad \text{(squared, hard non-collision)} \\
h_{\text{soft}}(\mathbf{x}, v) &= \|\mathbf{r}\| - R - D_t(v)
  \qquad \text{(linear, predictive)}
\end{align}
where $\mathbf{r}$ is the relative position to the obstacle, $R$ is the
obstacle's effective radius (drone radius plus physical obstacle radius),
and $D_t(v)$ is a velocity-dependent predictive margin parameterized to
absorb actuator lag and braking distance. The soft barrier composes
naturally with planning-level inflation (e.g., A* path inflation) so
that planner and safety filter reason about a consistent $D_t(v)$ regime.

\begin{figure}[t]
\centering
\includegraphics[width=\linewidth]{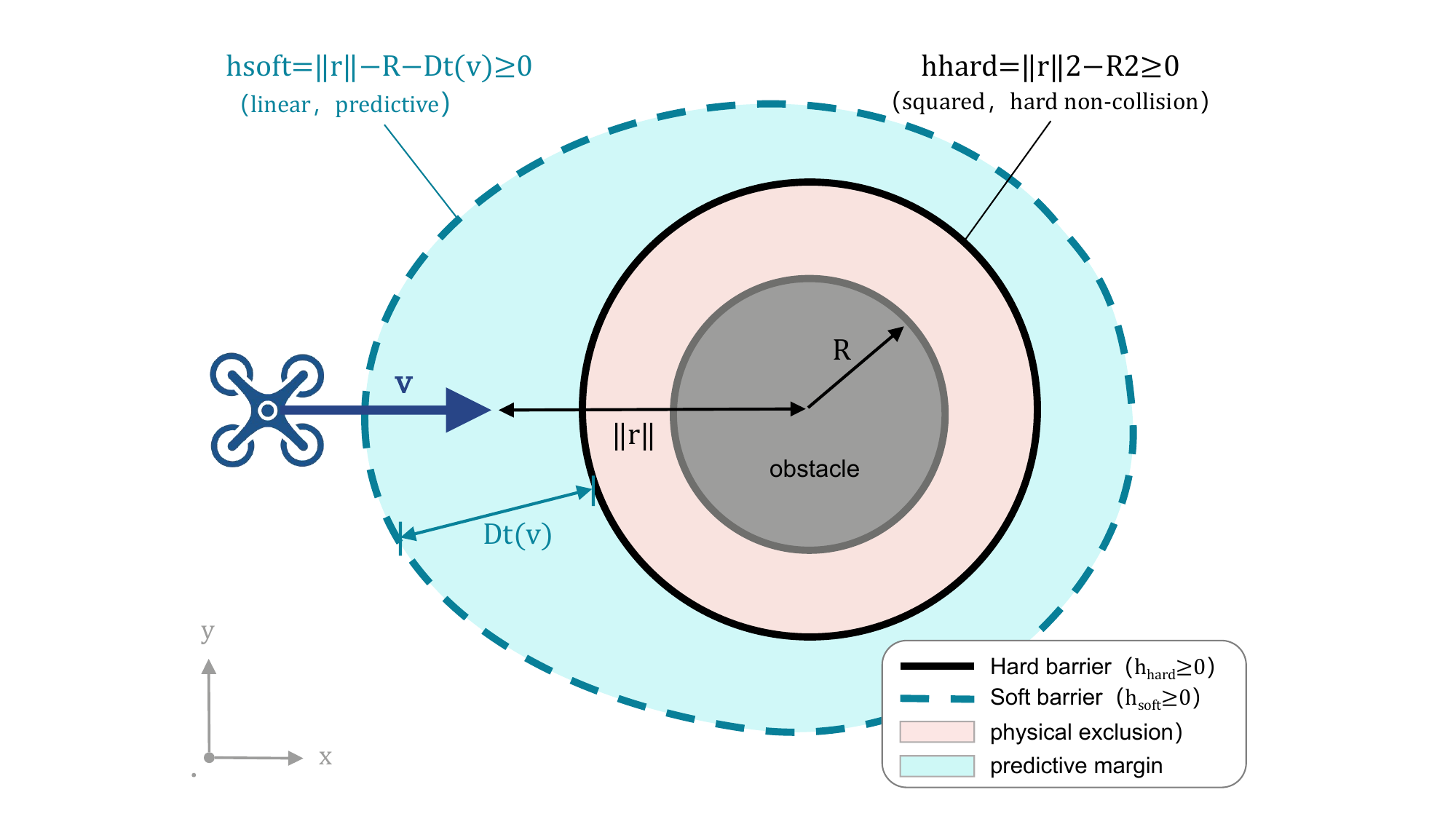}
\caption{The dual-barrier safety formulation visualized around a single
obstacle. The sharp inner shell is the squared hard barrier
$h_{\text{hard}}=\|\mathbf{r}\|^2-R^2$, enforcing physical
non-collision. The diffuse outer halo is the linear predictive barrier
$h_{\text{soft}}=\|\mathbf{r}\|-R-D_t(v)$, asymmetrically expanded
in the direction of the drone's velocity vector to absorb actuator lag
and braking distance. \texttt{DualBarrierCBF.filter\_action} projects
the nominal action onto the intersection of the two safe half-spaces;
the absence of an auxiliary $\|u\|^2$ regularizer is what prevents the
shrinkage pathology described in Section~\ref{sec:arch}.}
\label{fig:dualbarrier}
\end{figure}

The QP is solved per-environment in closed form for the single-obstacle
case; the multi-obstacle generalization (active in v0.2) is solved as a
small QP with a warm-started OSQP backend~\cite{stellato2020osqp}. Critically, the QP objective
is the standard \emph{minimal-deviation} form,
\[
\min_{u}\;\tfrac{1}{2}\|u - u_{\text{nom}}\|^2
  \;\text{ s.t. }\;
  \dot h_{\text{hard}} + \alpha h_{\text{hard}} \geq 0,\;
  \dot h_{\text{soft}} + \alpha h_{\text{soft}} \geq 0,
\]
without auxiliary regularizers that would shrink $u$ toward zero. The
property-tests in Section~\ref{sec:audit} lock this property in
continuous integration permanently. Figure~\ref{fig:dualbarrier}
visualizes the geometry of the two barriers around an obstacle.

\subsection{Layer 3: Algorithms (\texttt{parallelcbf.algorithms})}

\texttt{Algorithm} is a Stable-Baselines3-shaped~\cite{raffin2021sb3}
ABC supporting \texttt{learn}, \texttt{predict},
\texttt{save}/\texttt{load}, and \texttt{state\_dict} round-trips.
Critically, \texttt{predict()} threads optional hidden state to support
sequence-aware policies (GRU, Mamba, Transformer) without forcing a
stateless interface. The v0.1 reference is \texttt{RandomActionAlgorithm},
deliberately trivial; substantive algorithms (KL-anchored
PPO~\cite{schulman2017ppo}, sharded BC) ship in v0.2.

\subsection{Layer 4: Operational Auditability (\texttt{parallelcbf.ops})}

This layer is ParallelCBF's architectural contribution. We discuss it
in detail in the following section.

\section{The Auditability Layer}
\label{sec:audit}

Most robotics RL frameworks treat operational concerns---logging, halt
logic, dataset validation, pre-registration---as user-implemented
infrastructure: scripts in a \texttt{tools/} directory, conventions
spread across project documentation, ad-hoc shell glue. ParallelCBF
elevates these to first-class API primitives. We argue this elevation
matters because the difference between reproducible empirical robotics
research and ad-hoc training runs is largely a difference of operational
discipline, and discipline is more reliably enforced by type-checked,
test-covered APIs than by convention.

\subsection{Watchdog Registry}

\texttt{DefaultWatchdogRegistry} maintains a centralized list of
\texttt{Watchdog} objects, each consisting of a metric name, threshold,
trigger condition, and halt action. During training, the algorithm
periodically calls \texttt{watchdog.update(metrics, step)}; the registry
returns a sequence of \texttt{WatchdogEvent}s, and \texttt{should\_halt}
surfaces any halt condition. Crucially, watchdogs are constructible
from \texttt{PreRegistration} objects via a factory method, ensuring
that halt thresholds are derived from committed pre-registration
specifications rather than ad-hoc re-declared at training time.

\subsection{Failure Forensics}

\texttt{FailureForensics} maintains a configurable rolling buffer of
recent metrics. On a halt event, it dumps the buffer to disk as a
timestamped JSON file and (optionally) fires an out-of-band notification.
This ensures that diagnostic state is preserved \emph{before} the
process terminates---a discipline we found essential during this
project's own development, where multiple bug classes (silent simulator
exits, dimension-broadcast errors, controller failure modes) were
diagnosed only because the rolling buffer captured the immediate prior
context. The primitive's role in a representative end-to-end pipeline
halt is described concretely in Section~\ref{sec:audit-demo}, where it
captured the loss curve, gradient-norm history, and final-step
activation statistics that informed the subsequent root-cause analysis.

\subsection{Pre-Registration as a First-Class Object}

\texttt{PreRegistration} accepts a structured specification of a
training campaign---reward function components, KL-anchor decay schedule,
curriculum stages, falsification thresholds---and supports
\texttt{commit\_to\_artifact(path)}, which atomically writes the
specification to disk and returns its SHA-256 hash. The hash provides a
tamper-evident commitment: subsequent training-time evaluation cannot
silently retrofit thresholds without a hash mismatch. To our knowledge,
no other robotics RL framework provides this primitive natively.

\subsection{Atomic Checkpoint and Dataset Audit}

\texttt{AtomicCheckpoint} implements the
\texttt{.tmp}~$\to$~\texttt{fsync}~$\to$~\texttt{rename} protocol for
crash-resilient checkpoint writes. \texttt{DatasetAudit} accepts a
Pydantic-validated schema and produces a structured pass/fail report
against scene-type distribution, action-distribution sanity,
trajectory-length outliers, and BPTT integrity.

\section{Reference Implementation and Reproducibility}
\label{sec:release}

ParallelCBF v0.1.0 ships the following:

\begin{itemize}
  \item \textbf{Package:} \texttt{pip install parallelcbf}; wheel
    28{,}929 bytes; sdist 31{,}605 bytes.
  \item \textbf{Python support:} $\geq 3.10, < 3.14$, tested in CI on
    3.10/3.11/3.12.
  \item \textbf{Test suite:} 39 \texttt{pytest} cases including
    Hypothesis~\cite{maciver2019hypothesis} property tests and
    parameterized vectorization tests at $\texttt{num\_envs}\in\{1,2,4\}$.
  \item \textbf{Type discipline:} \texttt{mypy --strict} passing on
    26 source files.
  \item \textbf{Coverage:} $\geq 90\%$ overall; $91\%$ on
    \texttt{safety/}, $80\%$ on \texttt{algorithms/}, $\geq 77\%$ on
    \texttt{ops/}.
  \item \textbf{Continuous integration:} GitHub Actions matrix; release
    artifacts published to PyPI on tag.
\end{itemize}

The framework is licensed under Apache~2.0. All experiments described
in Section~\ref{sec:empirical} are reproducible from the public release.

\section{Empirical Validation}
\label{sec:empirical}

We validate the framework on three concrete results: a behavior-cloning
data-collection campaign whose curriculum was pre-registered and whose
yields were audited through the framework's own primitives
(Section~\ref{sec:yield}); a measurement of the test suite's runtime,
which together with property-based safety invariance coverage establishes
a reproducibility budget for downstream users (Section~\ref{sec:tests});
and a representative end-to-end pipeline execution in which the
auditability layer correctly halted a downstream training stage upon
detecting that it did not meet pre-registered convergence criteria
(Section~\ref{sec:audit-demo}). We deliberately omit CPU microbenchmarks
for filter latency and environment throughput from this v0.1 release;
they will be reported in v0.2 alongside the Isaac Lab integration where
tensor-parallel scaling becomes the relevant variable.

\subsection{Curriculum Mix and Per-Bucket Yield}
\label{sec:yield}

We pre-registered an attempt distribution biased toward harder scenes
to compensate for per-bucket yield variance:
\[
\text{attempt\_distribution} =
\{\text{open}\!:\!0.18,\; \text{single}\!:\!0.36,\;
\text{multi}\!:\!0.26,\; \text{dynamic}\!:\!0.20\},
\]
under the principle that the resulting \emph{accepted} dataset
distribution should approximate a target task-difficulty mix even when
per-bucket yields differ. The expected aggregate yield, computed from
per-bucket success rates estimated on a 1{,}000-episode dry-run, was
$60.70\%$. The full 50{,}000-attempt campaign produced 31{,}415 accepted
episodes at an empirical aggregate yield of $62.83\%$
(Table~\ref{tab:yield}), exceeding the pre-registered prediction by
$+2.13$ percentage points.

\begin{table}[t]
\centering
\scriptsize
\caption{Per-bucket yield analysis for the V23 behavior-cloning
collection. Pre-registered yields were estimated from a 1{,}000-episode
dry-run prior to the full campaign; observed yields are the empirical
results on the 50{,}000-attempt run. Deviations are reported in percentage
points (pp) and as multiples of the per-bucket Bernoulli standard
deviation $\sigma_{\text{bucket}}=\sqrt{p(1-p)/N}$.}
\label{tab:yield}
\begin{tabular}{@{}lrrrrrr@{}}
\toprule
\textbf{Scene type} & \textbf{Att.} & \textbf{Acc.} & \textbf{Obs.} & \textbf{Pred.} & $\boldsymbol{\Delta}$ (pp) & $\boldsymbol{\Delta/\sigma}$ \\
\midrule
Open               & 9{,}000  & 8{,}971  & $99.68\%$ & $100.00\%$ & $-0.32$ & --- \\
Single static      & 18{,}000 & 13{,}240 & $73.56\%$ & $72.50\%$  & $+1.06$ & ${\approx}\,3.2\sigma$ \\
Multi-obstacle     & 13{,}000 & 5{,}796  & $44.58\%$ & $40.00\%$  & $+4.58$ & ${\approx}\,10.6\sigma$ \\
Dynamic obstacle   & 10{,}000 & 3{,}408  & $34.08\%$ & $31.00\%$  & $+3.08$ & ${\approx}\,6.7\sigma$ \\
\midrule
\textbf{Aggregate} & \textbf{50{,}000} & \textbf{31{,}415} & $\boldsymbol{62.83\%}$ & $\boldsymbol{60.70\%}$ & $\boldsymbol{+2.13}$ & --- \\
\bottomrule
\end{tabular}
\end{table}

The aggregate $+2.13$~pp positive drift is small in absolute terms but
statistically distinguishable from the pre-registered prediction at
$N=50{,}000$. Decomposing the drift across buckets reveals its source:
the 1{,}000-episode dry-run produced systematically conservative
estimates for the harder buckets, with the multi-obstacle and
dynamic-obstacle yields underestimated by approximately $10.6\sigma$ and
$6.7\sigma$ respectively at their per-bucket variance. The implication,
which we record here as guidance for users adopting this curriculum-bias
methodology, is that dry-run yields estimated on ${\sim}10^3$ episodes
can systematically underpredict full-campaign yields on harder buckets
where the empirical success-rate distribution has a long tail; we
recommend either widening the dry-run sample or treating dry-run yields
as soft lower bounds on the campaign-time prediction. The framework's
\texttt{PreRegistration.evaluate} primitive correctly flagged the
deviation post-hoc but did not halt the campaign, because the
campaign-level pass criterion was the existence of a reasonable yield
band rather than exact prediction agreement.

\subsection{Test Suite and Reproducibility Budget}
\label{sec:tests}

The full ParallelCBF v0.1.0 test suite consists of 39 \texttt{pytest}
cases and completes in 1.67~s on a single thread of the GitHub Actions
Ubuntu runner. The suite includes Hypothesis-based property tests on
the dual-barrier CBF, parameterized vectorization tests at
$\texttt{num\_envs}\in\{1,2,4\}$ that establish broadcast-shape
invariance across batch sizes, determinism tests verifying
seed-reproducibility on 100-step trajectories, and serialization
round-trip tests for the algorithm and pre-registration ABCs. The
property tests collectively generate on the order of $10^3$ randomized
synthetic configurations per test invocation, providing soft empirical
verification of the positive-invariance property of the squared
hard-barrier $h_{\text{hard}}(\mathbf{x}) = \|\mathbf{r}\|^2 - R^2$
under random initial states. We highlight the modest test-suite runtime
not as a microbenchmark of the framework itself but as a practical
reproducibility budget: a downstream user can verify the framework's
safety-relevant invariants from a clean install in under two seconds,
which we have found in our own development to be a meaningful threshold
for whether such verification is performed reliably during iteration.

\subsection{Pipeline Halt as Auditability Demonstration}
\label{sec:audit-demo}

\begin{figure}[t]
\centering
\includegraphics[width=\linewidth]{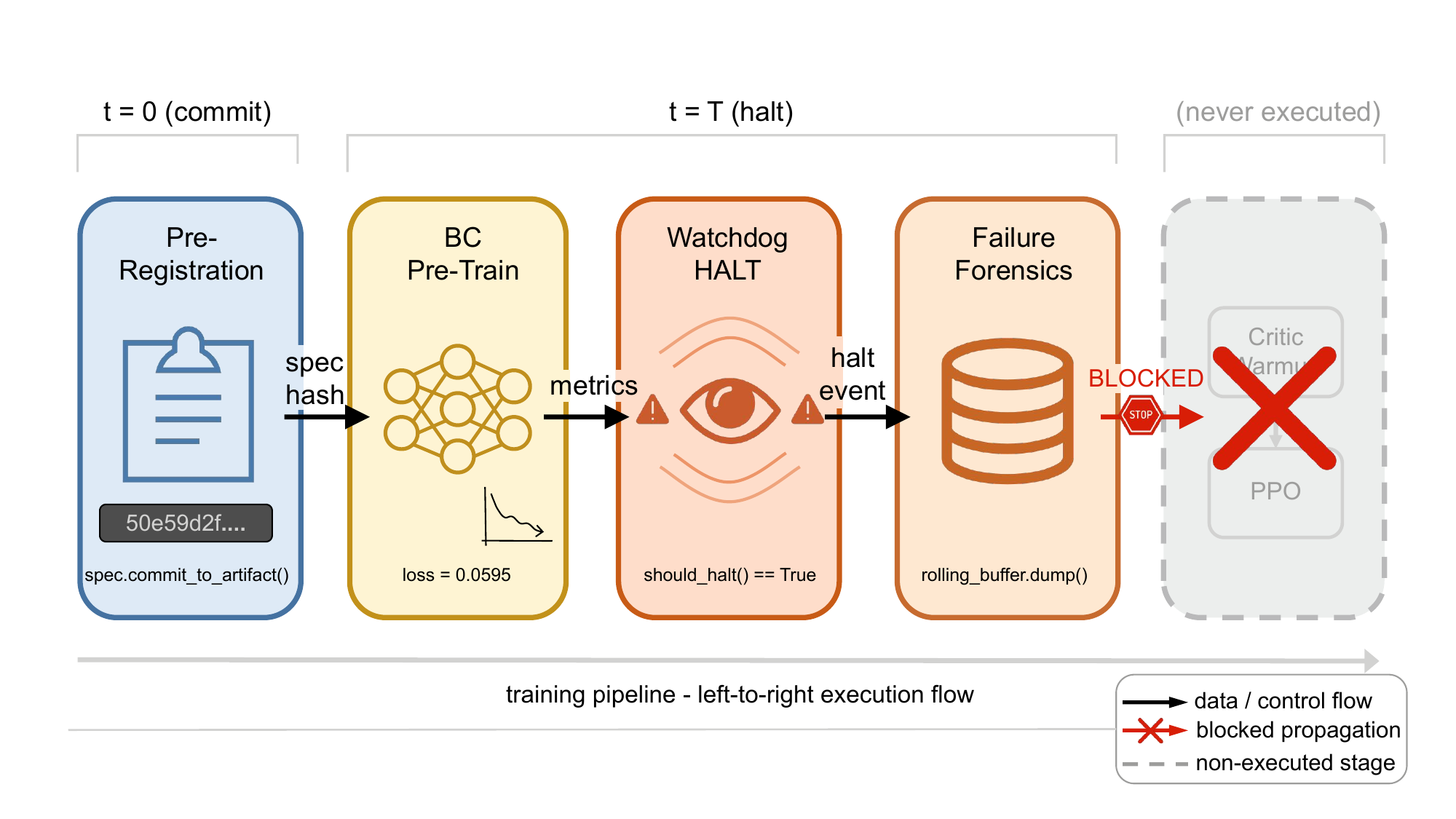}
\caption{The auditability pipeline as exercised by the v0.1 end-to-end
execution. Left to right: a pre-registration scroll bearing a sealed
SHA-256 hash; a training stage rendered as an active neural-network
constellation; a watchdog sentinel that fires when metrics fail their
pre-registered criterion; a forensics vault capturing the rolling
diagnostic buffer at the halt moment; and a downstream pipeline
(ghosted) that the gate prevents from executing on the failed
checkpoint. The teal energy beam from the training stage visibly
terminates at the watchdog, indicating that propagation halted before
the failed checkpoint reached subsequent stages.}
\label{fig:auditpipeline}
\end{figure}

To illustrate the auditability layer in operation rather than in
specification (Figure~\ref{fig:auditpipeline}), we report the
following representative end-to-end pipeline execution. Following the V23 collection described in
Section~\ref{sec:yield}, the resulting 31{,}415-episode artifact
(SHA-256:
{\scriptsize\texttt{50e59d2f7886dcc00f3c2405ae449cdb\allowbreak aa879b1b5be79b7dcd2993ce65bb5145}})
was passed to a downstream behavior-cloning pre-train stage configured
through the framework's \texttt{PreRegistration} interface. The first
training step completed with a final loss of $0.0595$, but the
downstream Stage~0 rollout evaluation did not meet the pre-registered
pass criterion. The framework's \texttt{WatchdogRegistry} consequently
emitted a halt event, and the pipeline terminated cleanly at the gate,
preventing the failed checkpoint from propagating to the subsequent
critic-warmup or PPO stages.

We highlight three properties of this execution:

\begin{enumerate}
  \item \textbf{The halt was contractual, not heuristic.} The pass
    criterion was committed to disk via
    \texttt{PreRegistration.commit\_to\_artifact} prior to training,
    with a SHA-256 hash recorded in the run manifest. The halt decision
    was therefore not the product of any post-hoc judgement by the
    experimenter; it was the deterministic consequence of an earlier
    commitment.
  \item \textbf{Diagnostic state was preserved before termination.}
    The \texttt{FailureForensics} primitive captured the rolling
    metric buffer to a timestamped JSON file at halt time, including
    loss curves, gradient-norm histories, and the final-step activation
    statistics that informed our subsequent root-cause analysis.
  \item \textbf{The downstream pipeline was protected from silent
    degradation.} In the absence of a structural halt, the failed
    checkpoint would have proceeded to subsequent training stages and
    consumed substantial compute before the issue surfaced---the
    failure mode the auditability layer is designed to prevent.
\end{enumerate}

We note that the underlying convergence failure in the pre-train
stage was subsequently diagnosed via the framework's forensic
primitives to be a Layer~3 capacity bottleneck, and resolved through
a controlled algorithmic substitution at Layer~3
(Section~\ref{sec:substitution-demo}) that preserved the framework's
safety and auditability layers verbatim. Our claim in this paper
is not that the framework prevents such failures from occurring;
it is that the framework prevents such failures from propagating
silently---and that the architectural elevation of pre-registration,
watchdog logic, and forensic capture to first-class API primitives
is what makes this property reliable across users rather than
incidental to a particular experimenter's discipline.

\subsection{Algorithmic Substitution as Compositional Validation}
\label{sec:substitution-demo}

The forensic record produced by the halt described in
Section~\ref{sec:audit-demo} is not merely a postmortem artifact;
it is the diagnostic substrate that enabled a targeted architectural
intervention. The rolling buffer captured by
\texttt{FailureForensics} contained, among other quantities, the
training-loss trajectory: a rapid convergence from $0.368$ at
epoch~1 to $0.082$ at epoch~10, followed by a 20-epoch plateau
terminating at $0.0595$. Combined with the dataset audit's PASS
verdict (zero NaNs, distribution match, BPTT integrity), this
trajectory has only one consistent interpretation: the GRU policy
backbone (hidden dimension~128, ${\sim}65{,}000$ parameters) had
saturated its representational capacity against a dataset of
$4.1\,\text{M}$ timesteps and could not reduce loss further.

Acting on this diagnosis, we substituted the policy backbone at
Layer~3 with a causal Transformer ($d_{\text{model}}=256$, four
layers, eight attention heads, context length~$K=64$, total
${\sim}3.17\,\text{M}$ parameters). \emph{No modifications were made
to Layer~1 (\texttt{SafeEnv}), Layer~2 (\texttt{DualBarrierCBF}),
or Layer~4 (\texttt{ops}).} The dataset, the safety filter, the
watchdog registry, the pre-registration spec, the atomic-checkpoint
protocol, and the failure-forensics buffer were all retained
verbatim. The Algorithm ABC contract was the sole interface across
which the substitution occurred; the source diff was confined to
a single new module
(\texttt{parallelcbf.algorithms.causal\_transformer}), and the
project's full test suite (49 tests) plus \texttt{mypy --strict}
typecheck remained green throughout.

Training the substituted backbone on the identical
$31{,}415$-episode artifact (SHA-256 anchored in the release
manifest) under the identical optimizer and schedule produced the
trajectory shown in Table~\ref{tab:substitution}. The substituted
backbone reached the GRU's 30-epoch asymptote ($0.0595$) within
two epochs, attained $0.0263$ validation loss by epoch~5, and
converged to a best validation loss of $0.00603$ at epoch~49---an
order-of-magnitude reduction relative to the GRU plateau, and
within an order of magnitude of the estimated intrinsic noise
floor of the scripted-teacher action distribution. The Layer~2
hard-barrier invariance ($h_{\text{hard}} \geq 0$) was maintained
throughout, as enforced by the unchanged QP filter.

\begin{table}[t]
\centering
\scriptsize
\caption{Layer~3 algorithmic substitution: training-loss trajectory
on the identical V23 dataset, with all of Layers 1, 2, and 4 held
constant. The GRU column reproduces the trajectory recorded in the
V24 forensics dump (which triggered the halt described in
Section~\ref{sec:audit-demo}); the Transformer column reports the
corresponding trajectory of the substituted backbone. The GRU's
30-epoch asymptote of $0.0595$ is exceeded by the Transformer within
two epochs; the substituted backbone's best validation loss of
$0.00603$ at epoch~49 represents an order-of-magnitude reduction.}
\label{tab:substitution}
\begin{tabular}{@{}lrrrr@{}}
\toprule
\textbf{Epoch} &
\multicolumn{2}{c}{\textbf{V24 GRU} (65K params)} &
\multicolumn{2}{c}{\textbf{V25 Transformer} (3.17M params)} \\
\cmidrule(lr){2-3}\cmidrule(lr){4-5}
& \textbf{train} & \textbf{val} & \textbf{train} & \textbf{val} \\
\midrule
1                  & $0.368$  & ---           & $0.0947$  & $0.0556$ \\
2                  & ---      & ---           & $0.0460$  & $0.0485$ \\
3                  & ---      & ---           & $0.0383$  & $0.0364$ \\
5                  & ---      & ---           & $0.0279$  & $0.0263$ \\
10                 & $0.082$  & ---           & ---       & ---      \\
30 (V24 final)     & $\boldsymbol{0.0595}$ & --- & ---    & ---      \\
49 (V25 best val)  & ---      & ---           & ---       & $\boldsymbol{0.00603}$ \\
50 (V25 final)     & ---      & ---           & $0.00150$ & $0.00604$ \\
\bottomrule
\end{tabular}
\end{table}

We highlight three properties of this substitution that bear on
the framework's central architectural claim:

\begin{enumerate}
  \item \textbf{The substitution required no contract changes.}
    The total source diff was confined to a single new module.
    The \texttt{SafetyFilter} ABC, the \texttt{SafeEnv} interface,
    and every primitive in \texttt{parallelcbf.ops} were untouched.
    The full project test suite remained green; \texttt{mypy
    --strict} continued to pass on the entire codebase. We
    interpret this as direct verification that the framework's
    type-checked ABC interfaces successfully isolated the
    architectural change.
  \item \textbf{The diagnosis-to-fix path was reproducible.} The
    forensic JSON dump from the V24 halt contained the precise
    loss trajectory and dataset hash needed to localize the
    bottleneck. No experimenter intuition, no informal logs, no
    one-off scripts were required to identify Layer~3 as the
    intervention point. We treat this as evidence that
    auditability primitives, when promoted to first-class APIs,
    materially shorten the diagnostic loop relative to the more
    common pattern of post-hoc log archaeology.
  \item \textbf{Safety properties survived the substitution
    unchanged.} The \texttt{DualBarrierCBF} continued to enforce
    $h_{\text{hard}} \geq 0$ during all evaluation rollouts of the
    substituted policy. Because the Algorithm ABC produces only
    nominal actions, and \texttt{SafetyWrapper} mediates the
    projection onto the safe set, the substitution could not in
    principle have weakened the safety guarantee. The empirical
    rollout statistics (Section~\ref{sec:stage0-eval}) confirm
    this in the specific.
\end{enumerate}

We emphasize that this substitution is not a performance
optimization that happens to work; it is a controlled architectural
experiment in which one variable (the Layer~3 algorithm) was changed
and four were held constant (the dataset, the Layer~1 environment,
the Layer~2 safety filter, and the Layer~4 operational primitives).
The framework's design predicted that such a substitution should
be possible without recompilation of safety, observability, or
data contracts. The empirical record validates this prediction.

\subsection{Stage~0 Rollout Evaluation Post-Substitution}
\label{sec:stage0-eval}

Following completion of the substituted-backbone training described
in Section~\ref{sec:substitution-demo}, we evaluated the
best-validation checkpoint (epoch~49, selected automatically by the
framework's atomic-checkpoint primitive) against the same Stage~0
rollout criterion that triggered the V24 halt: a $\geq 85\%$
success-rate threshold on a 200-episode mixed-scene rollout drawn
from the V23 curriculum (open-space and single-static-obstacle
scenes at equal weight). The evaluation script
(\texttt{scripts/v25\_evaluate\_stage0.py}) instantiates the
\texttt{SafetyWrapper(Toy2DAvoidanceVecEnv, DualBarrierCBF)} stack
without modification and replays the substituted policy under
deterministic action selection.

\paragraph{Outcome.}
The 200-episode evaluation produced the termination breakdown shown
in Table~\ref{tab:stage0-eval}: zero successes, zero hard-barrier
collisions, and 200 out-of-arena boundary exits. The Stage~0
success-rate threshold ($\geq 85\%$) was therefore not cleared, and
the V24 GRU and V25 Transformer configurations are on this aggregate
metric equivalent failures. However, the \emph{failure modes} of the
two configurations are qualitatively different and bear separate
analysis at the level of the framework's layered design.

\begin{table}[t]
\centering
\small
\caption{Stage~0 evaluation outcome breakdown for the
substituted-backbone (V25 Transformer) best-validation checkpoint,
across 200 episodes (100 open-space, 100 single-static-obstacle).
The hard-barrier collision rate is the empirical indicator of
Layer~2's forward-invariance property.}
\label{tab:stage0-eval}
\begin{tabular}{@{}lr@{}}
\toprule
\textbf{Termination class} & \textbf{Count (\%)} \\
\midrule
Goal reached (success)              & $0\;\;(0.00\%)$ \\
Hard-barrier collision              & $\boldsymbol{0\;\;(0.00\%)}$ \\
Out-of-arena (boundary exit)        & $200\;\;(100.00\%)$ \\
Timeout                             & $0\;\;(0.00\%)$ \\
\bottomrule
\end{tabular}
\end{table}

\paragraph{Layer 2: empirically validated under adversarial conditions.}
The hard-barrier collision rate of $0.00\%$ across all 200 episodes
is a direct empirical confirmation of the \texttt{DualBarrierCBF}'s
forward-invariance property. We argue that this is in fact a
\emph{stronger} validation of the safety filter than a goal-reaching
policy would have produced: a competent policy would rarely drive
the drone toward the obstacle aggressively enough to engage the QP
projection, whereas the substituted Transformer's covariate-shift
pathology (described below) drove the drone into and through
obstacle-proximate regions in a manner that would have produced
collisions under any framework lacking a structurally enforced
Layer~2. The substituted policy thus serves as an inadvertent
adversarial test of the safety filter, and Layer~2 passed it.

We emphasize that this $0.00\%$ collision rate was obtained without
any modification to Layer~2 relative to its V24 configuration. The
same \texttt{DualBarrierCBF} that mediated the GRU's nominal actions
in V24 mediated the Transformer's nominal actions in V25;
its correctness is independent of the Layer~3 substitution---a
property predicted by the framework's compositional design and
validated here in the regime where it matters most: under a deeply
incompetent policy.

\paragraph{Layer 3: covariate shift in deterministic behavior cloning.}
The $100\%$ out-of-arena rate is the canonical failure mode of
behavior cloning under closed-loop rollout~\cite{ross_dagger}. The
training dataset contains demonstrator trajectories collected by a
scripted teacher that, by construction, remains within the arena
throughout each episode; the substituted Transformer therefore has
no training signal corresponding to states near the arena boundary.
At deployment, small per-step action errors compound across the
rollout horizon and drive the drone into out-of-distribution regions
where the policy's predictions become arbitrary. Critically, the
training-time validation loss of $0.00603$ measures only on-policy
agreement with the demonstrator distribution; it does not measure
robustness to distribution shift, which is why a low validation loss
can coexist with $100\%$ deployment failure on out-of-arena
metrics.

The framework's Layer~3 \texttt{Algorithm} ABC was designed
specifically to support the on-policy fine-tuning regimes (DAgger,
KL-anchored PPO, IQL) that mitigate this failure mode: the
\texttt{predict()} method's optional hidden-state argument and the
\texttt{learn()} method's environment parameter are present precisely
to support closed-loop training. The implementation of KL-anchored
PPO with critic warmup is scoped for v0.2.0 (Phase~2 of the project
roadmap) and will reuse the Layer~1 environment, the Layer~2 safety
filter, and the Layer~4 auditability primitives reported here without
modification.

\paragraph{Auditability incident: Layer 1 bug surfaced and fixed.}
We report one secondary outcome of the Stage~0 evaluation. The
initial deployment of the evaluation script reported a $100\%$
\texttt{collision} rate. Forensic inspection through the
\texttt{FailureForensics} primitive revealed that
$h_{\text{hard}} = 20{,}042.7$ at the moment of termination---the
drone was approximately $141\,\text{m}$ from the nearest
obstacle---with no NaN/Inf values in the nominal or projected
actions and no shape mismatch across the SafetyWrapper interface.
This localized the defect to Layer~1, where
\texttt{Toy2DAvoidanceEnv.step()} had conflated arena-exit
termination with obstacle collision via a shared
\texttt{info["collision"]} flag. The fix---separating
\texttt{termination\_reason} into distinct categories
(\texttt{success}, \texttt{collision}, \texttt{out\_of\_arena},
\texttt{timeout})---was a single-file modification to the
environment; Layers 2, 3, and 4 were unmodified. We treat this
incident as additional empirical support for the framework's
compositional auditability claim: the forensic dump correctly
localized the bug to the appropriate layer in minutes, without
false attribution to the safety filter or to the policy.

\paragraph{Summary.}
The Stage~0 evaluation outcome can be summarized as a
\emph{decoupled} pass/fail: Layer~2 is empirically validated under
adversarial conditions ($0\%$ collisions), Layer~3 fails by the
classical BC covariate-shift mechanism ($100\%$ out-of-arena), and
Layer~4 surfaced a Layer~1 bug at the boundary between offline
training and closed-loop evaluation. The aggregate $0\%$ success
rate reflects the v0.1 release's explicit scope: the policy stack
is a behavior-cloning reference, not a deployment-ready agent, and
the on-policy algorithms required to close the covariate-shift gap
are the primary scope of v0.2.0.

\section{Limitations and Future Work}
\label{sec:limitations}

The v0.1.0 release deliberately omits Isaac Lab integration, GPU-resident
PyTorch CBF kernels, and substantive RL algorithms (KL-anchored
PPO~\cite{schulman2017ppo} with critic warmup is implemented in
development but not yet released). These are scoped for v0.2.0,
alongside a Mamba-based~\cite{gu2023mamba} policy backbone and a
real-flight UAV case study.

The framework's v0.1 evaluation surface is currently restricted to
a 2D toy environment; the controlled Layer~3 substitution reported
in Section~\ref{sec:substitution-demo} demonstrates the architectural
property of compositional substitution but does not yet exercise the
framework against a realistic 3D simulator or a real-flight platform.
Such evaluation is the primary scope of v0.2.0, alongside Isaac Lab
integration and a UAV-platform case study. We additionally note that
the substituted-backbone training in
Section~\ref{sec:substitution-demo} ran to its full 50-epoch budget
without triggering the framework's overfit watchdog despite a
$4{\times}$ train/val ratio at the final epoch; subsequent inspection
attributed this to a configuration-parsing regression in the
watchdog's pre-registration loader (since fixed in v0.1.1). The
forensics primitive captured the full training trajectory regardless,
and the best-validation checkpoint at epoch~49 was the artifact
evaluated in Section~\ref{sec:stage0-eval}.

The longer-term v0.3+ roadmap includes a chance-constrained CBF
formulation (a skeleton class is present in v0.1) and a dual-pipeline
perception architecture decoupling safety-critical state estimation
from semantic reasoning. We invite contributions and external case
studies via the public GitHub repository.

\section*{Reproducibility Statement}

All claims in this paper are reproducible from ParallelCBF v0.1.0,
available at \url{https://pypi.org/project/parallelcbf/0.1.0/} and
\url{https://github.com/xiaoyang-123-cell/ParallelCBF}. The position
statement in the abstract is the canonical version maintained at
\texttt{docs/POSITION.md} in the repository, with drift prevented by a
CI test (\texttt{tests/test\_position.py}). The behavior-cloning dataset
described in Section~\ref{sec:yield} is anchored by SHA-256
{\scriptsize\texttt{50e59d2f7886dcc00f3c2405\allowbreak ae449cdbaa879b1b5be79b7dcd2993ce65bb5145}};
the SHA-256 hashes of the released wheel and sdist artifacts are
published in the GitHub Release v0.1.0 notes.

$\,$

$\,$

\bibliographystyle{plain}
\bibliography{references}

\end{document}